\documentclass[10pt,twocolumn,letterpaper]{article}

\usepackage[pagenumbers]{cvpr} 

\usepackage{amsmath, amssymb, algorithm, algpseudocode, caption}

\usepackage{multirow}
\usepackage{soul}

\usepackage[dvipsnames]{xcolor}

\newcommand{\boldstart}[1]{\vspace{3pt}\noindent\textbf{#1}}

\usepackage{listings}
\usepackage{xcolor}

\definecolor{keywordcolor}{rgb}{0.3,0.3,0.7}
\definecolor{commentcolor}{rgb}{0.5,0.5,0.5}
\definecolor{stringcolor}{rgb}{0.7,0.3,0.3}
\definecolor{backgroundcolor}{rgb}{0.95,0.95,0.95}

\definecolor{cvprblue}{rgb}{0.21,0.49,0.74}
\usepackage[pagebackref,breaklinks,colorlinks,citecolor=cvprblue]{hyperref}

\def\paperID{67} 
\def\confName{3DV\xspace}
\def\confYear{2025\xspace}

\title{CatFree3D: Category-agnostic 3D Object Detection with Diffusion}

\author{
Wenjing Bian\textsuperscript{1}\quad
Zirui Wang\textsuperscript{1}\quad
Andrea Vedaldi\textsuperscript{2} \\
\textsuperscript{1}Active Vision Lab \quad \textsuperscript{2}Visual Geometry Group \\
University of Oxford\\
{\tt\small \{wenjing, ryan, vedaldi\}@robots.ox.ac.uk} \\
\url{https://bianwenjing.github.io/CatFree3D}}

\begin{document}
\maketitle
\begin{abstract}
Image-based 3D object detection is widely employed in applications such as autonomous vehicles and robotics, yet current systems struggle with generalisation due to complex problem setup and limited training data. 
We introduce a novel pipeline that decouples 3D detection from 2D detection and depth prediction, using a diffusion-based approach to improve accuracy and support category-agnostic detection. 
Additionally, we introduce the Normalised Hungarian Distance (NHD) metric for an accurate evaluation of 3D detection results, addressing the limitations of traditional IoU and GIoU metrics. 
Experimental results demonstrate that our method achieves state-of-the-art accuracy and strong generalisation across various object categories and datasets.
\end{abstract}

\section{Introduction}%
\label{sec: intro}

Image-based 3D object detection systems are designed to identify and localise objects in three-dimensional space from input images. These systems play a critical role in applications like autonomous vehicles and robotics.

Recent developments in deep learning have significantly advanced 3D object detection~\cite{chen2016monocular, simonelli2019disentangling, zhang2021objects, nie2020total3dunderstanding, brazil2023omni3d, li2024unimode, rukhovich2022imvoxelnet, mousavian20173d}.
While these methods have been well engineered, they remain largely domain-specific and are limited in the number of detectable object categories, especially when compared to state-of-the-art 2D detection systems~\cite{carion2020end, reis2023real}, which can detect hundreds of categories across various domains.

The performance gap between 2D and 3D detection, particularly in generalising to more detectable categories, is mostly due to i) the complex problem setup and ii) insufficient training data.
The 3D detection task is closely related to various other tasks such as 2D detection, depth estimation and object pose estimation, each of which is a challenging research area.
Additionally, labelling 3D data is more labour-intensive, requiring specifying nine degrees of freedom instead of four in 2D.
This combination of complexity and limited data restricts current 3D detection methods to fewer categories, often with reduced accuracy.

To overcome these limitations, we propose a pipeline that decouples the 3D detection task from 2D detection and depth prediction.
This decoupling enhances training efficiency, and, most importantly, allows for a category-agnostic approach with improved accuracy.

Our key idea lies in recovering a 3D bounding box from a random noise, conditioned on several visual prompts, using a denoising network inspired by diffusion models~\cite{ho2020denoising, song2020denoising} in a generative fashion.
Specifically, the random noise is sampled from a normal distribution, with visual prompts consisting of the image of the target object, a 2D detection bounding box, and the depth of the target.
During training, we take advantage of ground truth labels of 2D bounding boxes and object depths.
During inference, the model can be integrated with 2D detectors and depth estimation models or take prompts from various sources, \eg human annotation.

In addition to simplifying existing pipelines and achieving category-agnostic detection, our diffusion-based approach allows us to generate an \textit{arbitrary} number of predictions for a \textit{single} target due to their stochastic nature.
We leverage this by estimating multiple 3D bounding boxes for one target, assigning a confidence score to each, and selecting the most confident one, further improving detection accuracy.

While developing our novel 3D detection method, we discovered that conventional metrics, such as Intersection over Union (IoU) and Generalised IoU (GIoU), often struggle to evaluate 3D detection results accurately, particularly for non-overlapping or enclosing cases, which are common for thin and small objects.
To address these limitations, we propose a new metric called \emph{Normalised Hungarian Distance} (NHD), which seeks a one-to-one assignment between the corners of the ground truth and predicted 3D bounding boxes, then calculates the Euclidean distance between corresponding corners, offering a more detailed and precise evaluation of 3D object detection.

In summary, we make three key contributions:
\ul{First}, we introduce a novel diffusion-based pipeline for 3D object detection that decouples the 3D detection task from 2D detection and depth prediction, enabling category-agnostic 3D detection.
\ul{Second}, we enhance the 3D detection accuracy by leveraging generative capabilities in our diffusion-based pipeline to predict multiple bounding boxes with confidence scores.
\ul{Third}, we propose the \emph{Normalised Hungarian Distance} (NHD), a new evaluation metric that provides a more precise assessment of 3D detection results.

As a result, our method achieves state-of-the-art accuracy in 3D object detection in a category-agnostic manner and demonstrates strong generalisation to unseen datasets.

\section{Related Work}

\boldstart{2D Object Detection.}
2D object detectors include two-stage detectors~\cite{girshick2015fast, ren2015faster} that use a coarse-to-fine approach and single-stage detectors~\cite{carion2020end, reis2023real, ren2015faster, liu2016ssd} that directly estimate the location of objects from the extracted visual features.
DiffusionDet~\cite{chen2023diffusiondet} was the first to apply diffusion to detection tasks, progressively refining noisy 2D boxes towards the target objects.
Category-agnostic 2D detection models are conceptually similar to our approach in that they avoid using category information.
These models learn to differentiate generic objects from the image background using low-level visual cues~\cite{zitnick2014edge, o2015learning} or using supervision from bounding box labels~\cite{maaz2022class, jaiswal2021class, kim2022learning}.
However, instead of learning objectness like these methods do, we focus on mapping 2D boxes to 3D boxes by leveraging visual cues from the input image.

\boldstart{Monocular 3D Object Detection}
Monocular 3D object detectors predict 3D cuboids from a single input image.
Depending on the dataset domain, certain models are tailored for outdoor self-driving scenes~\cite{chen2016monocular, huang2022monodtr, ranasinghe2024monodiff, liu2021ground, ma2019accurate, mousavian20173d, reading2021categorical, wang2022probabilistic, mousavian20173d}, while others are specifically designed for indoor environments~\cite{nie2020total3dunderstanding, huang2018cooperative, tulsiani2018factoring, kulkarni20193d}.
Additionally, some studies~\cite{rukhovich2022imvoxelnet, brazil2023omni3d, li2024unimode} have explored integrating both indoor and outdoor datasets during training.
These methods often use category labels for supervision~\cite{brazil2023omni3d, huang2022monodtr, li2024unimode, liu2021ground}, require category information as a prior for initialisation or as input~\cite{nie2020total3dunderstanding, brazil2023omni3d, ranasinghe2024monodiff}, or focus on specific scenes and object categories with strong assumptions about the predictions' dimensions or orientation~\cite{simonelli2019disentangling, yan2024monocd, chen2016monocular, mousavian20173d}.
This reliance on category and scenario-specific knowledge limits their generalisation to in-the-wild scenes and novel categories.
In contrast, our approach does not use category information during training or inference, focusing solely on predicting 3D bounding boxes.
This allows the model to be used for novel objects that were not present during training.

\boldstart{Diffusion Models for Visual Perception}
Diffusion models~\cite{ho2020denoising, sohl2015deep, song2019generative, song2020score} have demonstrated remarkable results in computer vision~\cite{harvey2022flexible, yang2023diffusion, ho2022video}, natural language processing~\cite{austin2021structured, gong2023diffuseq, li2022diffusion, yu2022latent}, and multimodal data generation~\cite{avrahami2022blended, ramesh2022hierarchical, rombach2021high, poole2022dreamfusion, zhang2023adding}.
In visual perception tasks, DiffusionDet~\cite{chen2023diffusiondet} was the first to apply box diffusion for 2D object detection from a single RGB image. 
Additionally, diffusion has been utilised for tasks like image segmentation~\cite{amit2021segdiff, chen2023generalist} and human pose estimation~\cite{gong2023diffpose, holmquist2023diffpose}.
For 3D object detection, Zhou \emph{et al.}~\cite{zhou2023diffusion} introduced diffusion of 5 DoF Bird's Eye View boxes as proposals for detection from point clouds.
Diffusion-SS3D~\cite{ho2024diffusion} employs diffusion for semi-supervised object detection in point clouds, denoising object size and class labels.
DiffRef3D~\cite{kim2023diffref3d} and DiffuBox~\cite{chen2024diffubox} apply diffusion to refine proposals/coarse boxes for 3D object detection from point clouds.
MonoDiff~\cite{ranasinghe2024monodiff} uses Gaussian Mixture Models to initialise the dimensions and poses of 3D bounding boxes, recovering the target boxes through diffusion conditioned on an image.
Unlike these approaches, which assume an initial distribution or proposals for the diffusing box parameters~\cite{ranasinghe2024monodiff, chen2024diffubox}, limited DoFs for box orientation and dimensions~\cite{kim2023diffref3d, chen2024diffubox}, or diffuse only partial parameters~\cite{ho2024diffusion, ranasinghe2024monodiff}, our model initialise all box parameters with random noise for diffusion.
By conditioning the model on an image and a 2D box, our model recovers using diffusion the in-plane translations, three sizes, and three DoF for the rotation of the 3D box.
\begin{figure*}[t]
\centering
\includegraphics[width=0.8\textwidth]{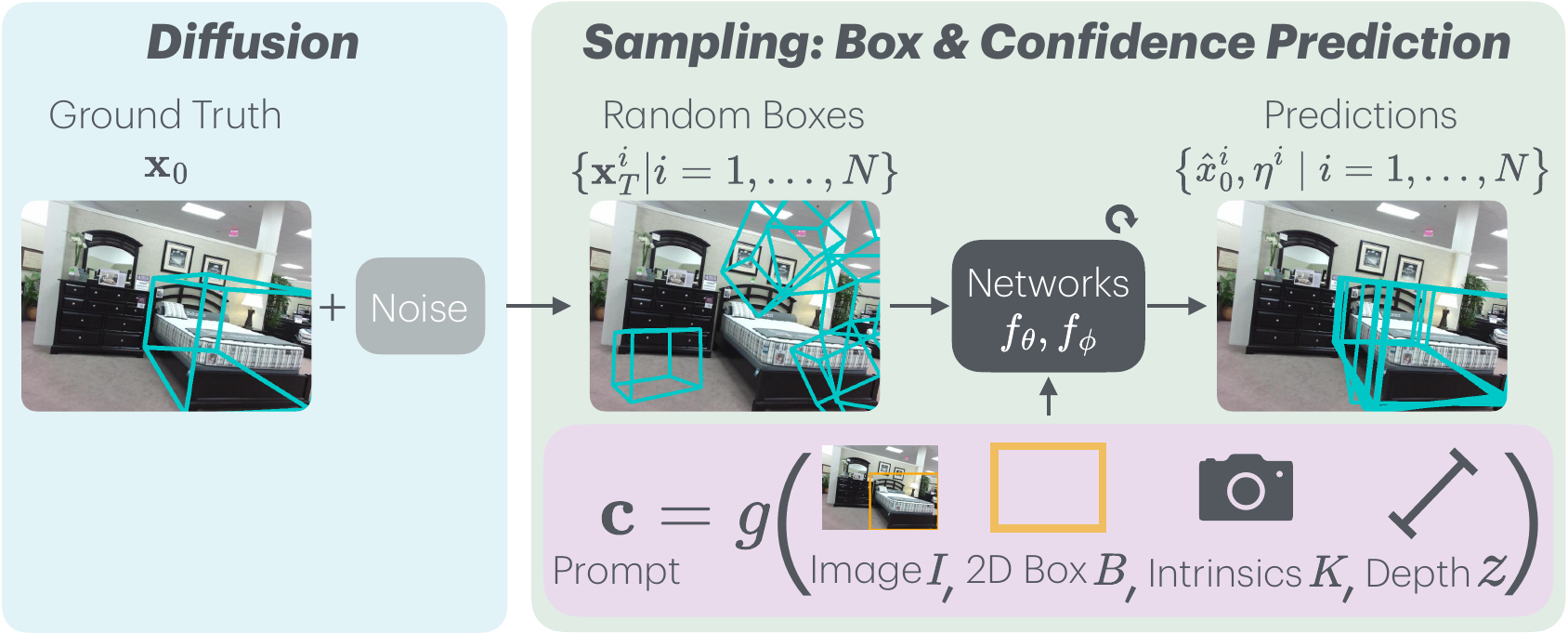}
\caption{\textbf{Method Overview.} 
During forward diffusion, we add $N$ independent Gaussian noises to a ground truth box $\mathbf{x}_0$ to obtain a number of noisy boxes. We then train a denoising network $f_\theta$ to recover the target box parameters $\hat{\mathbf{x}}_0$ from noisy boxes, conditioned on a vision-related signal $\mathbf{c}$. Additionally, we train another network $f_\phi$ to estimate a confidence score $\eta$ for each predicted box. The final output is the box with the highest confidence score.
}\label{fig:method:overview}
\end{figure*}

\section{Method}%
\label{sec:method}

Given an image $I$, a 2D bounding box of an object $B$, the object depth $z$, and the camera intrinsics $K$, our objective is to estimate the centre, 3D size and orientation of a 3D box that tightly encloses the object. 
We formulate this task as a conditional diffusion process~\cite{ho2020denoising}, progressively recovering a target box from a noise sampled from normal distributions, conditioned on multiple prompts.

Let us consider a general diffusion setup first.
In the forward diffusion process, Gaussian noise is incrementally added to a variable $\mathbf{x}_0$ over $T$ time steps until it follows a normal distribution. 
During backward denoising, an estimate $\hat{\mathbf{x}}_0$ can be recovered from its noisy version $\mathbf{x}_t$ using a neural network $f_\theta$:
\begin{equation}
    f_\theta(\mathbf{x}_t, t)  \rightarrow \hat{\mathbf{x}}_0,
    \label{eq:general_diffusion}
\end{equation}
where $t \in [1, T]$ denotes a diffusion step. 
\textit{In our 3D object detection task, we consider $\mathbf{x}_0$ the parameters of a 3D box}.

To adapt the general diffusion process to this vision-based 3D detection setup, we consider a conditional denoising network $f_\theta$:
\begin{equation}
\label{eq: 2}
    f_\theta(\mathbf{x}_t, t, \mathbf{c}) \rightarrow \hat{\mathbf{x}}_0,
\end{equation}
where
$\mathbf{c}$ denotes the conditional signal that includes information from the input image $I$, the 2D bounding box $B$, the camera intrinsics $K$, and the object depth $z$. 
This conditional diffusion process is similar to diffusion-based text-to-image generation tasks~\cite{rombach2021high, zhang2023adding}, where the conditional signal typically consists of text descriptions.

In \cref{sec:method:preparation} we begin by introducing the parametrisation of the 3D bounding boxes and the prompt encoding.
We then explain the forward diffusion and reverse sampling in \cref{sec:method:diffusion} and \cref{sec:method:sampling} respectively, detailing how we utilise the generative properties of diffusion networks to predict multiple proposals and their associated confidence scores.
Finally, we outline our training processes and losses in \cref{sec:training}.


\subsection{Preparation}%
\label{sec:method:preparation}

\boldstart{Box Parameterisation}
We consider the position, orientation and size of a 3D bounding box.
Specifically, each 3D bounding box $\mathbf{x}_0$ is represented using 11 parameters:
\begin{equation}
    \mathbf{x} := [u, v, w, h, l, \mathbf{p}].
\end{equation}
The position of the 3D box is defined by the 2D projected coordinates $u$ and $v$, which represent the object centre on the image plane of the observing camera.
This position parametrisation decouples the image plane position components and the depth component, allowing us to leverage object depth from various sources.
The orientation of the object relative to the camera of the input image is represented by the continuous 6D allocentric rotation $\mathbf{p} \in \mathbb{R}^6$~\cite{zhou2019continuity}. 
The size of the 3D bounding box is captured by $w$, $h$, and $l$.
Overall, this parametrisation follows~\cite{brazil2023omni3d}, but with the depth component excluded.

\boldstart{Prompts Encoding}
Our conditioning signal $\mathbf{c}$ is derived from the image $I$, a 2D bounding box $B$, camera intrinsics $K$, and the object depth $z$:
\begin{equation}
\mathbf{c} = g(I, B, K, z),
\end{equation}
where function $g(\cdot)$ denotes a prompt encoding function. 
This function includes an image encoding backbone, positional encoding functions, and a shallow MLP that summaries all prompt information in preparation for the box prediction network.
Further details on the prompt encoding can be found in the supplementary material.

\begin{figure*}[t]
    \centering
    \includegraphics[width=\textwidth]{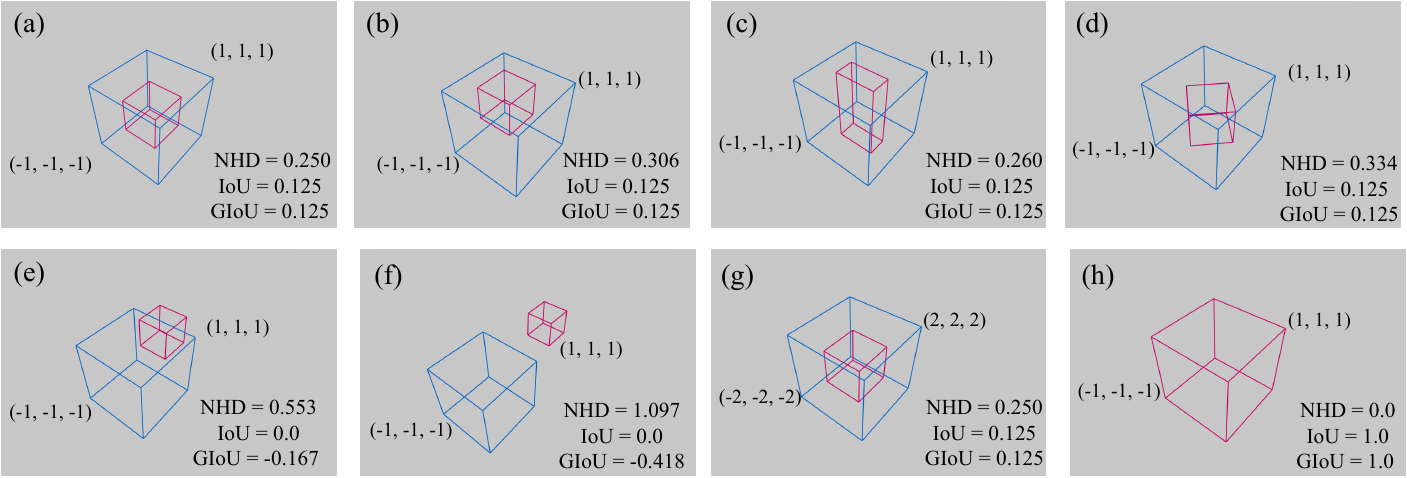}
    \caption{\textbf{Comparison Between NHD, IoU and GIoU}. Comparing block (a) with blocks (b, c, d, e, f), we show that NHD provides a more accurate measurement of errors compared to IoU and GIoU, particularly under translation, scaling, and rotation transformations. Block (g) demonstrates that all three metrics are scale-invariant. Block (h) presents metric values when the two boxes are perfectly aligned.}\label{fig: simulate}
\end{figure*}

\subsection{Diffusion: Adding Noise to a Box}%
\label{sec:method:diffusion}

In the forward diffusion process, we start with a noise-free 3D bounding box, denoted as $\mathbf{x}_0$, and iteratively add Gaussian noise over $T$ steps to generate a fully noisy box $\mathbf{x}_T$, which follows a normal distribution.
This process follows the standard DDPM schedule~\cite{ho2020denoising}.

\boldstart{Preprocessing}
Before applying noise to the original 3D bounding box $\mathbf{x}_0$, we perform additional normalisation and scaling to ensure that $\mathbf{x}_0$ lies within the range $[-s, s]$, where $s$ corresponds to the signal-to-noise ratio of the diffusion process.
During the \ul{normalisation} step, the projected coordinates $u$ and $v$ are normalised relative to the image dimensions.
The dimensions of the 3D bounding box—width $w$, height $h$, and length $l$—are normalised against a predefined maximum box size.
The orientation of the box $\mathbf{p}$, expressed in the allocentric representation~\cite{zhou2019continuity}, is inherently normalised and does not require further adjustment.
In the \ul{scaling} step, all box parameters are further scaled by a scalar $s$.
As shown in \cref{sec: discussion}, this scaling step, adopted from~\cite{chen2023generalist, chen2023diffusiondet}, improves box prediction accuracy.

\subsection{Sampling: Predicting a Box}%
\label{sec:method:sampling}

Our method predicts a 3D bounding box for a target object using a denoising network $f_\theta$, conditioned on a vision-related prompt $\mathbf{c}$.
To enhance detection performance, we introduce a confidence score for each predicted box, generating multiple candidate boxes and selecting the one with the highest confidence.

\boldstart{Single Box Prediction}
To predict a single 3D box $\hat{\mathbf{x}}_0$, we begin with a randomly sampled noise $\mathbf{x}_T$ and iteratively refine the 3D box $\hat{\mathbf{x}}_t$ with our denoising network $f_\theta$, conditioned on the encoded prompt $\mathbf{c}$.
This process continues until the final denoising step $t = 0$, following the standard sampling procedure introduced in DDIM~\cite{song2020denoising}.

\boldstart{Multi-Box Prediction and Selection}
Leveraging the generative capabilities of diffusion networks, our method allows for the prediction of \textit{multiple} 3D boxes for a \textit{single} target object.
Specifically, for each target object, we sample $N$ 3D box parameters $\{\mathbf{x}_T^i | i= 1, \dots, N\}$ from a normal distribution, producing $N$ predictions $\{\hat{\mathbf{x}}_0^i | i= 1, \dots, N\}$. 
We introduce a learnable confidence score $\eta^i$ for each prediction, where we select the box with the highest confidence score as the final prediction during inference.

\boldstart{Confidence Prediction}
To estimate the uncertainty  $\mu \in (0, \infty)$ for each box prediction, we employ an additional network branch $f_\phi$, which takes the current box estimation and the vision conditioning signal $\mathbf{c}$ as inputs:
\begin{equation}
f_\phi(\mathbf{c}, \mathbf{x}_t) \rightarrow \mu.
\end{equation}
The confidence score $\eta \in (0, 1)$ is derived from the uncertainty through an exponential mapping $\eta = e^{-\mu}$. This score reflects the agreement between the current box estimation and the provided vision prompt.



 

\subsection{Training}%
\label{sec:training}

We follow the standard DDPM~\cite{ho2020denoising} training process and utilise a training loss $\mathcal{L}$ consisting of two loss terms: 
\begin{equation}
    \mathcal{L} = \mathcal{L}_{\text{3D}} + \lambda_{\text{reg}} \mathcal{L}_{\text{reg}},
\end{equation}
with $\lambda_{\text{reg}}$ being a hyperparameter that balances the regularisation term $\mathcal{L}_{\text{reg}}$ and the reconstruction term $\mathcal{L}_{\text{3D}}$. 

The reconstruction loss $\mathcal{L}_{\text{3D}}$ is designed to encourage our denoising network to predict accurate 3D boxes.
This is achieved by penalising the Chamfer distance between the corners of predicted boxes and the ground truth, weighted by the confidence $\eta$:
\begin{equation}
\label{eq: diffusion2}
    \mathcal{L}_{\text{3D}} = \frac{1}{N} \sum_{i=1}^N \eta^i \mathcal{L}_{\text{chamfer}}(\hat{\mathbf{x}}_0^i, \mathbf{x}_0), 
\end{equation}
whereas the regularisation term $\mathcal{L}_{\text{reg}}$ prevents the uncertainty prediction $\mu$ being excessively large:
\begin{equation}
    \mathcal{L}_{\text{reg}} = \frac{1}{N} \sum_{i=1}^N \mu^i;    
\end{equation}
Details regarding the Chamfer distance $\mathcal{L}_{\text{chamfer}}$ between two 3D boxes are provided in the supplementary material.

\begin{figure}[t]
    \centering
    \includegraphics[width=1.0\columnwidth]{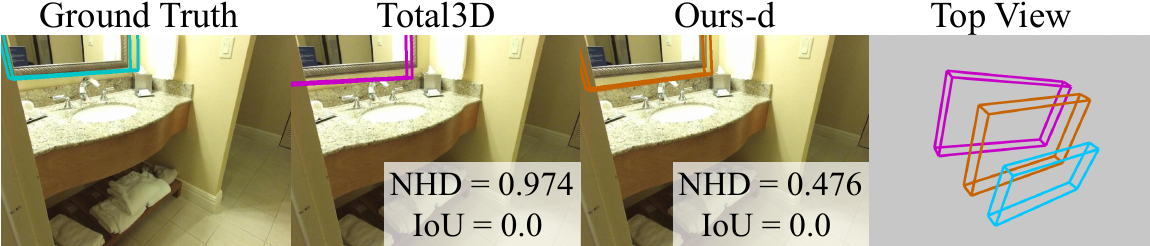}
    \caption{\textbf{IoU and NHD in a Practical Example.} For thin objects like mirrors, even a small translational offset can lead to an IoU of 0. In contrast, NHD effectively captures and reflects the box estimation error in these cases.}%
    \label{fig: metric}
\end{figure}

\begin{table*}[t]
\caption{\textbf{Detection Performance: Comparing Our Method with Cube R-CNN~\cite{brazil2023omni3d} and Total3D~\cite{nie2020total3dunderstanding} on Omni3D Dataset.} 
The top three rows use GT 2D boxes along with predicted depths. The depths of our predictions are set to the same as \cite{brazil2023omni3d} for fair comparison. The bottom three rows use GT 2D boxes and GT depths for all methods.
}
\label{table: main}
\centering
\begin{tabular}{lccccccccc}
\hline
\multirow{2}{*}{Methods} &  & \multicolumn{2}{c}{SUN RGB-D}  &  & \multicolumn{2}{c}{Omni3D Indoor} &  & \multicolumn{2}{c}{Omni3D Indoor \& Outdoor}     \\ \cline{3-4} \cline{6-7} \cline{9-10} 
                         &  & IoU (\%) $\uparrow$ & NHD  $\downarrow$            &  & IoU (\%) $\uparrow$ & NHD  $\downarrow$            &  & IoU (\%) $\uparrow$ & NHD  $\downarrow$            \\ \hline
Total3D                  &  & 24.8          & 0.376          &  & -             & -              &  & -             & -              \\
Cube R-CNN               &  & 36.2          & 0.236          &  & 19.5          & 0.667          &  & 23.0          & 0.593          \\
Ours-d                   &  & \textbf{40.2} & \textbf{0.231} &  & \textbf{20.8} & \textbf{0.648} &  & \textbf{23.3} & \textbf{0.591} \\ \hline
Total3D*                 &  & 46.6          & 0.184          &  & -             & -              &  & -             & -              \\
Cube R-CNN*              &  & 54.5          & 0.137          &  & 41.0          & 0.189          &  & 45.8          & 0.167          \\
Ours*                    &  & \textbf{61.4} & \textbf{0.114} &  & \textbf{49.7} & \textbf{0.143} &  & \textbf{51.4} & \textbf{0.142} \\ \hline
\end{tabular}
\end{table*}

\begin{table*}[t]
\caption{\textbf{Generalisation Performance: Novel Categories.} We train the model on 31 object categories from the SUN RGB-D training set and evaluate its IoU on 7 unseen categories. To show the importance of estimating in-plane offsets and box orientation, we use a Unprojection baseline that converts GT 2D boxes to 3D with GT depth and dimensions, setting 3D rotation to zero degrees.}
\centering
\begin{tabular}{l|l|lllllll|l}
Methods   & Trained on & sofa          & table         & cabinet       & toilet        & bathtub       & door          & oven          & avg.         \\
\hline
Unprojection & N/A        & 28.2          & 27.1          & 28.6          & 25.6          & 23.9          & 23.6          & 37.2          & 27.7         \\
Ours      & SUN RGB-D  & \textbf{56.4} & \textbf{56.8} & \textbf{53.0} & \textbf{61.3} & \textbf{46.7} & \textbf{27.7} & \textbf{62.1} & \textbf{52.0}
\end{tabular}%
\label{table: unseen}
\end{table*}

\section{Metric: Normalised Hungarian Distance}%
\label{sec: model_metric}

Intersection-over-Union (IoU) is a common metric for evaluating 3D object detection~\cite{brazil2023omni3d, rukhovich2022imvoxelnet, nie2020total3dunderstanding}.
While IoU is scale-invariant, it fails to measure the closeness of predictions when two boxes do not overlap, which is problematic for thin and small objects like mirrors or televisions.
An example is shown in \cref{fig: metric}.
Later, Generalised IoU (GIoU)~\cite{rezatofighi2019generalized} has been proposed to address this, but it still does not fully capture alignment in terms of centre, scale, and orientation, as seen in \cref{fig: simulate}.

We propose a new metric, \emph{Normalised Hungarian Distance} (NHD), to provide a more precise evaluation for 3D object detection.
NHD is calculated as 
\begin{equation}
\text{NHD}(\mathcal{M}_\text{pred}, \mathcal{M}_\text{gt}) = \frac{1}{d_{\text{gt}}} \sum_{i} \|a_i - b_{j}\|_2,
\end{equation}
where $P$ represents the optimal 1-to-1 mapping between predicted box corners $\mathcal{M}_\text{pred}$ and ground truth box corners $\mathcal{M}_\text{gt}$.
The mapping $P$ is obtained through a linear assignment algorithm by minimising the Euclidean distance between corresponding corners, ensuring corner $a_i \in \mathcal{M}_\text{pred}$ in the prediction matches corner $b_j \in \mathcal{M}_\text{gt}$ in the ground truth.
To make NHD scale-invariant, we normalise it with the maximum diagonal length $d_{\text{gt}}$ of the ground truth box.

\begin{figure}[t]
    \centering
    \includegraphics[width=0.95\columnwidth]{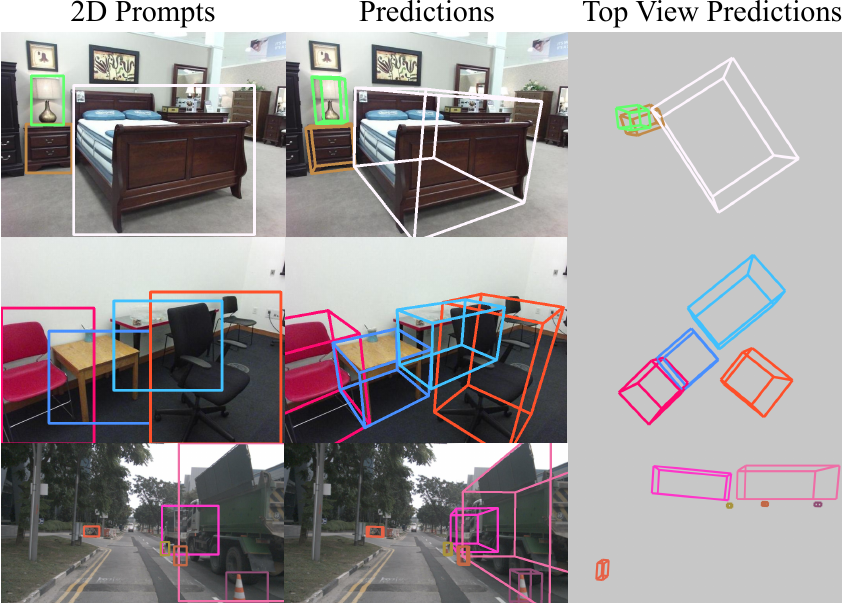}
    \caption{\textbf{Detection Performance: Results on Omni3D Test Set}. Estimating 3D box from GT 2D boxes and GT object depths.}%
    \label{fig: main1}
\end{figure}

\section{Experiments}

We outline our experimental setup in \cref{sec: implementation}.
In \cref{sec: main_results,sec: unseen}, we compare our approach to baselines, demonstrating its accuracy and generalisation.
We also present its application in 3D dataset labelling (\cref{sec: application}) and discuss the impact of hyperparameter choices (\cref{sec: discussion}).

\begin{figure*}[t]
    \centering
    \includegraphics[width=0.92\linewidth]{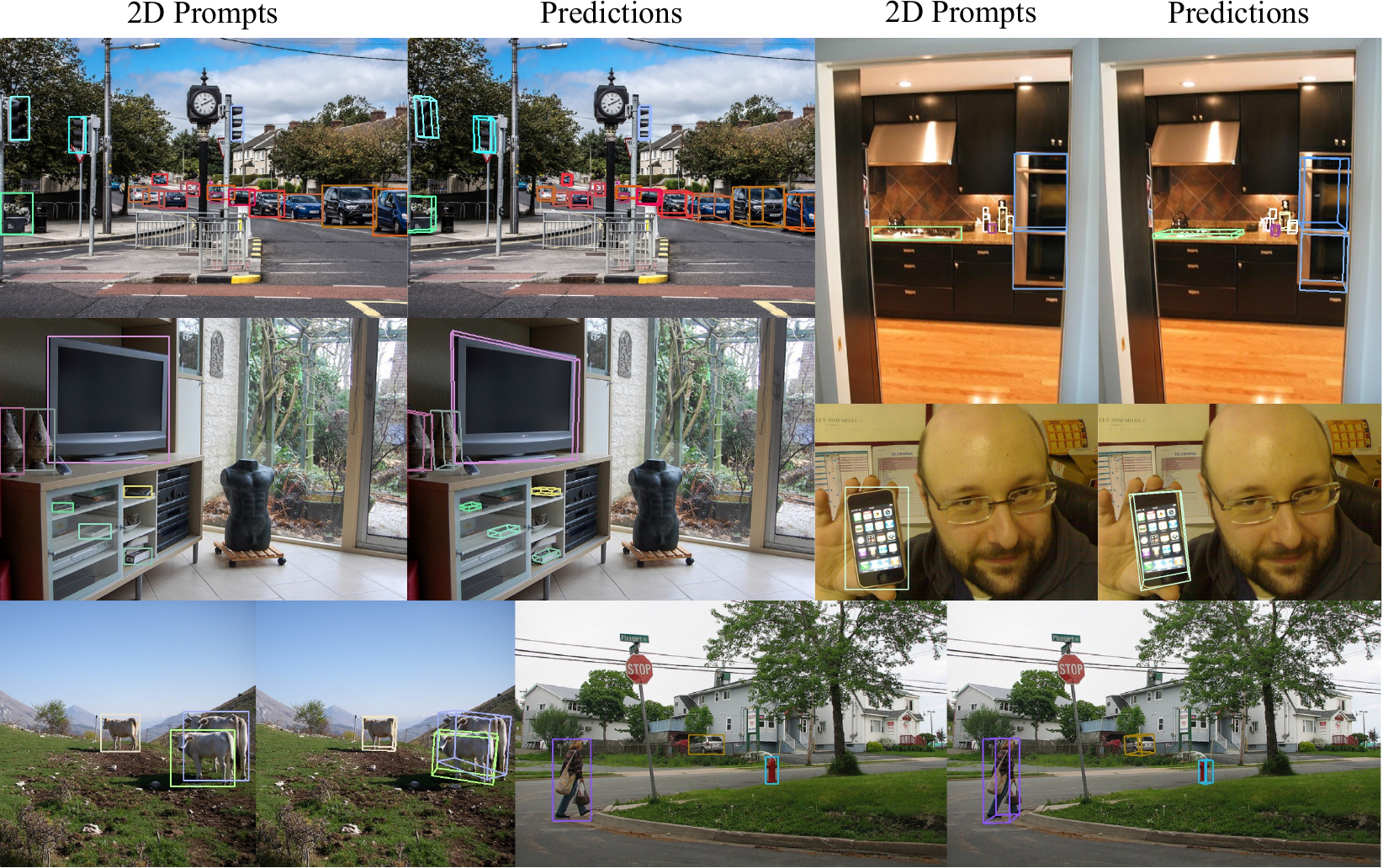}
    \caption{\textbf{Generalisation Performance: Results for In-the-Wild Objects on COCO Dataset.} 
    We show predictions made by our method without knowing object depths or camera intrinsics. By using constant values for depths and camera intrinsics, our approach accurately predicts 3D boxes with well-aligned projections on the image.}%
    \label{fig: coco}
\end{figure*}

\subsection{Experimental Setup}%
\label{sec: implementation}

\boldstart{Datasets}
For training and evaluation, we utilise the Omni3D dataset~\cite{brazil2023omni3d}, which is composed of six datasets\cite{song2015sun}, ARKitScenes~\cite{baruch2021arkitscenes}, Hypersim~\cite{roberts2021hypersim}, Objectron~\cite{ahmadyan2021objectron}, KITTI~\cite{geiger2012we}, and nuScenes~\cite{caesar2020nuscenes}, encompassing both indoor and outdoor environments.
For the quantitative tests in \cref{sec: application}, we employ the COCO~\cite{lin2014microsoft} and nerfstudio~\cite{nerfstudio} datasets.

\boldstart{Baselines}
To accurately assess the performance of our 3D detector, it is essential to eliminate errors from sources such as 2D detection and depth estimation.
In \cref{sec: main_results}, we assume ground truth 2D boxes and depth are provided for all methods during evaluation.
We select Cube R-CNN~\cite{brazil2023omni3d} and Total3DUnderstanding~\cite{nie2020total3dunderstanding} as our primary baselines, as their 3D detection heads can function independently.
Additionally, since the baseline methods require category information, we provide ground truth category labels to their models.

\boldstart{Metrics}
In \cref{sec: main_results}, we use IoU and NHD as two metrics to evaluate the performance of 3D detectors.

\boldstart{Model}
Our model architecture is built on Detectron2~\cite{wu2019detectron2}.
We use the Swin Transformer~\cite{liu2021swin} pre-trained on ImageNet-22K~\cite{deng2009imagenet} as the image feature encoder and freeze all parameters during training.
The denoising decoder adopts the iterative architecture in Sparse R-CNN~\cite{sun2021sparse} with 6 blocks, where the predictions from the previous stage are used as the input for the next stage.
Predictions from each stage are used to compute loss against the ground truth.

\boldstart{Training \& Inference}
We train the model for 270k iterations with a batch size of 16 on 2 A5000 GPUs.
In comparison, Cube R-CNN is trained on 48 V100 GPUs with a batch size of 192.
We use AdamW optimiser~\cite{kingma2014adam} with a learning rate of $2.5 \times 10^{-5}$ which decays by a factor of 10 at 150k and 200k iterations.
During training, we use image augmentation including random horizontal flipping and resizing.
During evaluation, as the box parameters are initialised with random noise, the result can vary with different random seeds.
To obtain a stable result, experiment results reported in \cref{sec: main_results} and \cref{sec: unseen} are computed by averaging the results of 10 different random seeds.
We also analyse the stability of model performance with random seeds in the supplementary material.

\begin{figure*}[t]
    \centering
    \includegraphics[width=0.8\linewidth]{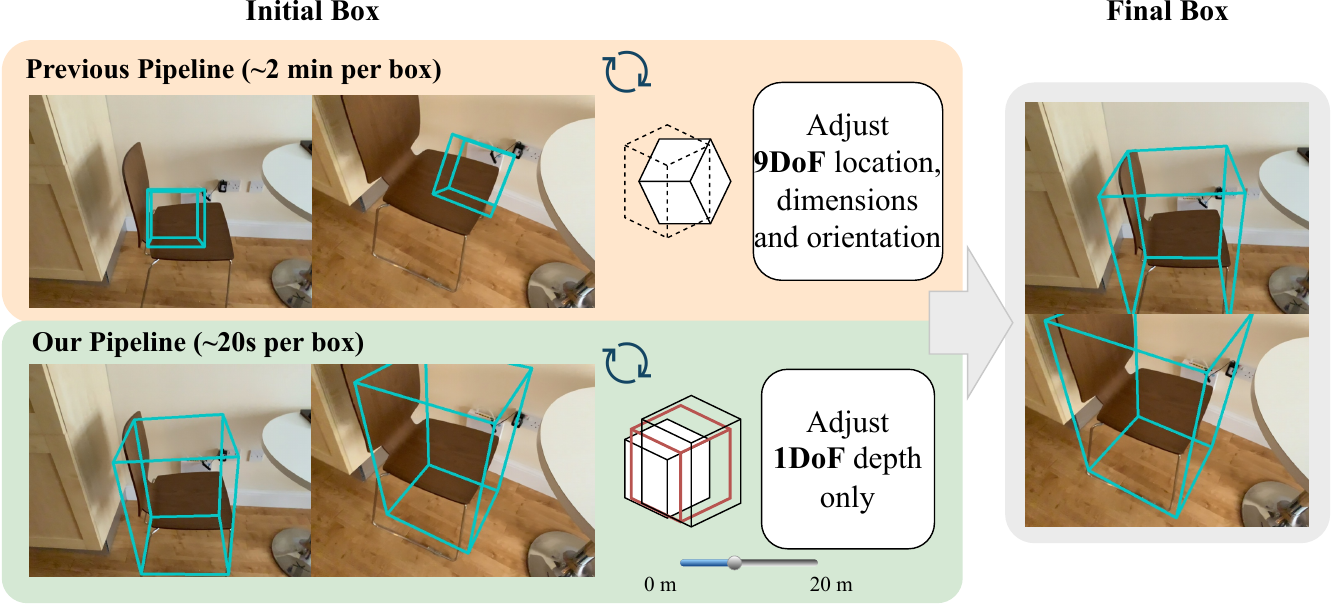}
    \caption{\textbf{Application: 3D Detection Dataset Annotation.} In a conventional 3D box annotation pipeline, annotators typically need to adjust a randomly initialised 3D box across nine degrees of freedom (rotation, translation, and size) until it appears correct in every view. This process is time-consuming and labour-intensive. Our model streamlines this workflow by reducing the task to a single degree of freedom, depth, significantly accelerating the dataset labelling process.}
    \label{fig: annotation}
\end{figure*}

\subsection{Detection Performance}%
\label{sec: main_results}

We compare our model with other 3D object detection approaches on Omni3D dataset~\cite{brazil2023omni3d} and its subsets.
For baseline models~\cite{brazil2023omni3d, nie2020total3dunderstanding}, we assume ground truth 2D boxes as the oracle 2D detection results and provide category information.
In contrast, our method requires 2D boxes only and does not use the category information.
As our model requires object depth as input, we use the depths estimated by~\cite{brazil2023omni3d} for a fair comparison.
We introduce a variant of each method that substitutes estimated object depths with ground truth depths.
As shown in \cref{table: main}, our approach outperforms the baselines, even with their use of additional category information and dimension priors.
The comparison between using predicted depths (upper half of table) and ground truth depths (lower half) highlights that inaccurate monocular depth estimation is a key source of error in monocular 3D detection models.
\cref{fig: main1} shows the qualitative performance of our model on the Omni3D test set.

\subsection{Generalisation Performance}%
\label{sec: unseen}

\begin{figure}[b]
    \centering
    \includegraphics[width=0.95\columnwidth]{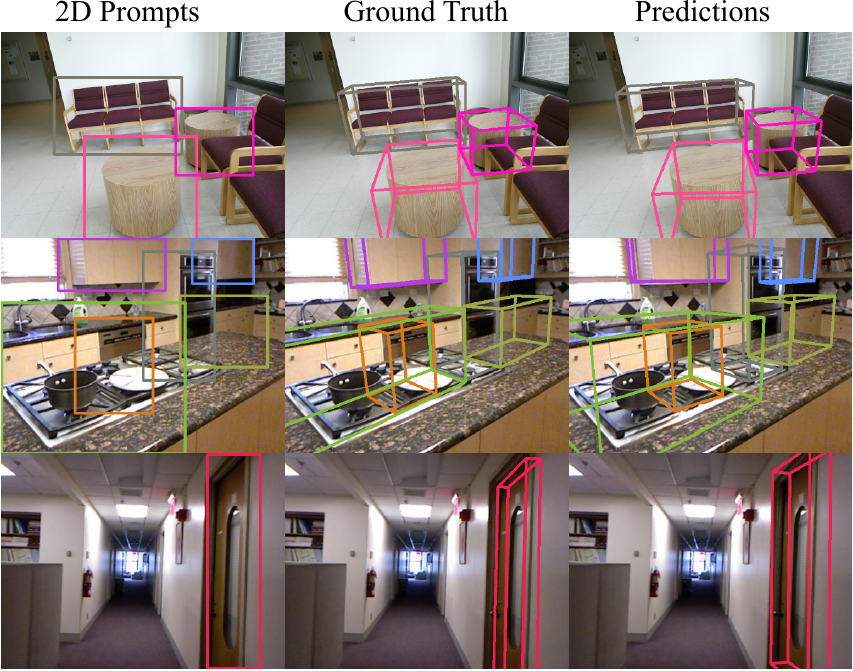}
    \caption{\textbf{Generalisation Performance: Predictions on Novel Categories}. 
    We show that our model generalises well to unseen object categories.
    }%
    \label{fig: unseen}
\end{figure}

\boldstart{Cross-category Generalisation}
To verify our model's generalisation ability across different objects, we trained it on the SUN RGB-D dataset using 31 out of 38 categories and evaluated it on the remaining categories.
Since there are no previous baselines for comparison, we created a simple baseline using axis-aligned 3D boxes to demonstrate the importance of estimating in-plane offsets and box orientation, as well as the effectiveness of our approach.
This baseline method takes the 2D box prompt and back-projects it to 3D using the provided depth and focal length.
The bounding box orientation is set to match the camera orientation, and the dimension in the z-direction is set to the maximum length of the ground truth bounding box in that direction.
Results in \cref{table: unseen} and \cref{fig: unseen} show that our approach significantly outperforms this baseline and demonstrates good generalisation ability.

\boldstart{Cross-dataset Generalisation}
To assess the model's generalisation across different datasets and input types, we perform a qualitative analysis with Omni3D-trained models on test images from COCO and nerfstudio datasets.

\Cref{fig: coco} shows the predictions on COCO, using only 2D prompts and uniform depth and camera intrinsics for all objects.
Despite inaccuracies in depth and intrinsics, our model accurately estimates up-to-scale 3D boxes where the projections align with the objects in the image.

\begin{figure}[b]
    \centering
    \includegraphics[width=1.0\columnwidth]{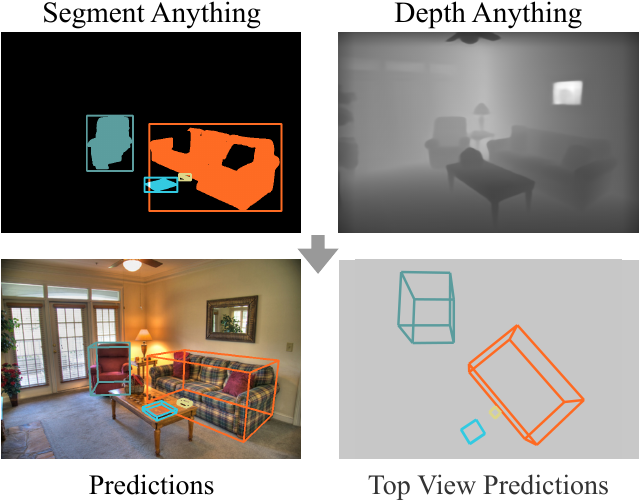}
    \caption{\textbf{Generalisation Performance: Predict with Prompts from 2D Detectors and Monocular Depth Estimators.}
    We infer object depths using DepthAnything~\cite{depthanything} and SegmentAnything~\cite{kirillov2023segment} for COCO Dataset.}%
    \label{fig: monodepth}
\end{figure}

\Cref{fig: monodepth} demonstrates the use of a monocular depth estimation network~\cite{depthanything} and a segmentation network~\cite{kirillov2023segment}, combined with 2D prompts, to estimate 3D boxes with accurate top-view projections. The depth for each object is calculated by averaging the depths within the object's mask.

\begin{figure}[]
    \centering
    \includegraphics[width=1.0\columnwidth]{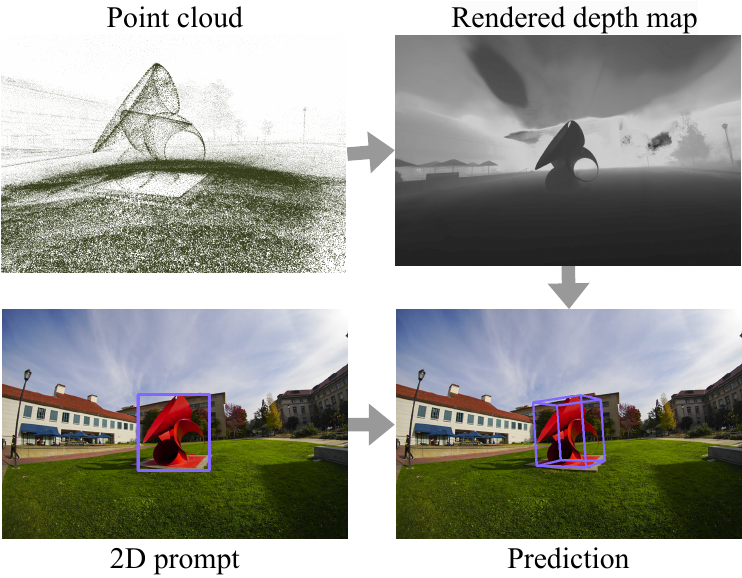}
    \caption{\textbf{Generalisation Performance: Predict with Prompts from human annotated 2D bounding boxes and a point cloud.} 
    }%
    \label{fig: pcd}
\end{figure}

When 3D data, such as point clouds or CAD models, is available, it can also be utilised for depth inference.
\Cref{fig: pcd} illustrates an example from nerfstudio dataset where 3D point clouds and a manually annotated 2D prompt are used to estimate the 3D box.

\begin{table*}[t]
\caption{\textbf{Discussion: Model Analysis on SUN RGB-D Test Set.}}\label{table: snr}
\begin{subtable}[t]{0.3\textwidth}
\caption{IoU and NHD under various SNR ($s$).}
\centering
\begin{tabular}{l|ll}
SNR & IoU (\%) $\uparrow$ & NHD $\downarrow$   \\ \hline
1.0 & 57.2  & 0.195   \\
2.0 & \textbf{59.8} & \textbf{0.177} \\
3.0 & 59.7  & 0.178
\end{tabular}
\end{subtable}%
\hfill
\begin{subtable}[t]{0.3\textwidth}
\caption{IoU and NHD under various DDIM steps.}\label{table: steps}
\centering
\begin{tabular}{l|ll}
Iter steps  & IoU (\%) $\uparrow$ & NHD $\downarrow$   \\ \hline
1          & 61.19  & 0.1153 \\
3          & 61.23  & 0.1150 \\
5          & \textbf{61.28}  & \textbf{0.1147}
\end{tabular}
\end{subtable}
\hfill
\begin{subtable}[t]{0.3\textwidth}
\caption{IoU under various No\@. of sampled boxes.}\label{table: props}
\centering
\begin{tabular}{l|ccc}
$N_\text{train}$ \textbackslash $N_\text{eval}$ & 1        & 10       & 100      \\ \hline
1           & 58.1     & 60.2     & 60.2     \\
10          & 60.9     & 61.2     & 61.2     \\
100         & 61.2     & \textbf{61.4}     & \textbf{61.4}
\end{tabular}
\end{subtable}
\end{table*}

\subsection{Application}%
\label{sec: application}




As discussed in \cref{sec: intro}, the availability of 3D detection datasets is significantly more limited compared to 2D detection, primarily due to the high costs associated with annotating 3D bounding boxes.
\Cref{fig: annotation} illustrates a typical 3D box annotation process, where annotators must draw a 3D box and fine-tune its projections across multiple views.
This manual adjustment of the box's location, dimensions, and orientation can take several minutes per box.

Our model streamlines this process by reducing the complexity from adjusting nine degrees of freedom (DoF) to just one DoF -- depth.
Starting with a 2D prompt from an annotator, our model generates a 3D box prediction with depth ambiguity.
The human annotator only needs to adjust the predicted box depth and verify it across other views.
Incorporating this approach into the 3D box annotation pipeline has the potential to greatly improve annotation efficiency.

\subsection{Discussion}%
\label{sec: discussion}

We conducted a model analysis on the 10 common categories in the SUN RGB-D test set to study its performance.
Unless otherwise stated, all inferences are run with 10 predictions for each object at a single sampling step.

\boldstart{Signal-to-noise Ratio}
In \cref{table: snr}, we show the influence of the diffusion process's signal-to-noise ratio (SNR).
Setting SNR to 2 achieves the highest performance in IoU and NHD, which is consistent with the observations in other diffusion models for detection~\cite{chen2023diffusiondet, kim2023diffref3d}.

\boldstart{Inference Iteration Steps}
During inference, we observed that increasing the number of DDIM sampling steps enhances model performance, albeit at the cost of longer inference times.
As shown in \cref{table: steps}, raising the iteration steps from 1 to 5 results in performance gains, which plateau with additional steps.

\boldstart{Number of Sampled Boxes.}
Since we use random noise to initialise box parameters, the number of boxes sampled is flexible and can vary between training ($N_\text{train}$) and inference ($N_\text{eval}$).
\Cref{table: props} presents the results for different combinations of $N_\text{train}$ and $N_\text{eval}$.
We observed that increasing $N_\text{train}$ enhances model performance, with a cost of larger memory consumption.
However, increasing $N_\text{eval}$ beyond 10 does not yield additional improvement.
Consequently, we select $N_\text{train} = 100$ and $N_\text{eval} = 10$ for 
other experiments.




\section{Conclusion}%
\label{sec:conclusions}
Our diffusion-based pipeline significantly improves 3D object detection by decoupling it from 2D detection and depth prediction, enabling category-agnostic detection.
Furthermore, the introduction of the Normalised Hungarian Distance (NHD) metric addresses the limitations of existing evaluation methods, providing a more accurate assessment of 3D detection outcomes, especially for complex scenarios involving small or thin objects. Experimental results confirm that our method achieves state-of-the-art accuracy and strong generalisation across various object categories and datasets.

{
    \small
    \bibliographystyle{ieeenat_fullname}
    \bibliography{main}
}
\clearpage
\appendix
\section{Implementation Details}
\subsection{Prompt Encoding}
We encode the image $I$, a 2D bounding box $B$, camera intrinsics $K$, and the object depth $z$ into the conditioning signal $\mathbf{c}$ through
\begin{equation}
\mathbf{c} = g(I, B, K, z),
\end{equation}

The 2D bounding box $B$ is described by the centre of the box along with its height and width on the image plane, i.e. 
\begin{equation}
    B := [u_\text{2d}, v_\text{2d}, w_\text{2d}, h_\text{2d}]
\end{equation}
To account for variation in depth and focal length, we further unproject the width and height of the 2D box into 3D using the following equation:
\begin{equation}
    (w_\text{3d}, h_\text{3d}) = (w_\text{2d} \frac{z}{f_{x}}, h_\text{2d} \frac{z}{f_{y}}), 
\end{equation}
where $f_x$ and $f_y$ are the focal lengths from the intrinsics $K$.

For the input image $I$,
we first encode it with a pre-trained Swin Transformer~\cite{liu2021swin} to generate multi-scale feature maps $F$. Next,
we extract local image features inside the region of the 2D box prompt to obtain $F_\text{RoI}$.   Additionally, we apply a cross-attention layer~\cite{vaswani2017attention} between $F$ and the 2D box $B$ to obtain $F_\text{atten}$.

By concatenating the transformed box prompt, image features and the object depth, the final conditioning signal $\mathbf{c}$ can be written as
\begin{equation}
\mathbf{c} = [F_{\text{RoI}}, F_{\text{attn}}, u_\text{2d}, v_\text{2d}, w_\text{3d}, h_\text{3d}, z].
\end{equation}
\subsection{Loss Function}
The Chamfer distance between the corners of the predicted 3D boxes $\mathcal{M}_\text{pred} = \{a_i| i= 1 ... 8\}$ and the corners of the ground truth boxes $\mathcal{M}_\text{gt} =\{b_i| i= 1 ... 8\}$ is computed as
\begin{equation}
\label{eq: cd}
    \mathcal{L}_\text{chamfer} = \sum_{a_i \in \mathcal{M}_\text{pred} }\min_{b_i \in \mathcal{M}_\text{gt}} \|a_i - b_i\|_1 + \sum_{b_i \in \mathcal{M}_\text{gt}} \min_{a_i \in \mathcal{M}_\text{pred}} \|b_i - a_i\|_1.
\end{equation}
\subsection{Baseline Models}
\begin{figure*}[t]
    \centering
    \includegraphics[width=0.9\linewidth]{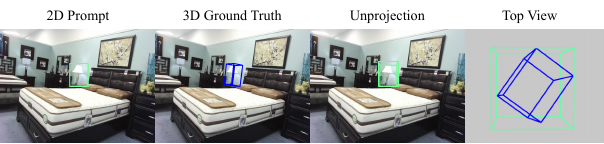}
    \caption{\textbf{Unprojection Baseline Illustration.} The Unprojection baseline (green) converts GT 2D boxes to 3D using GT depth and dimensions that match the GT 3D box (blue), with the 3D rotation to zero degrees. }%
    \label{fig: unproject}
\end{figure*}

\boldstart{Unprojection}
\cref{fig: unproject} illustrates how we obtain the Unprojection baseline for experiments in Sec.5.3 of the main paper.

\begin{figure}[t]
    \centering
    \includegraphics[width=1.0\linewidth]{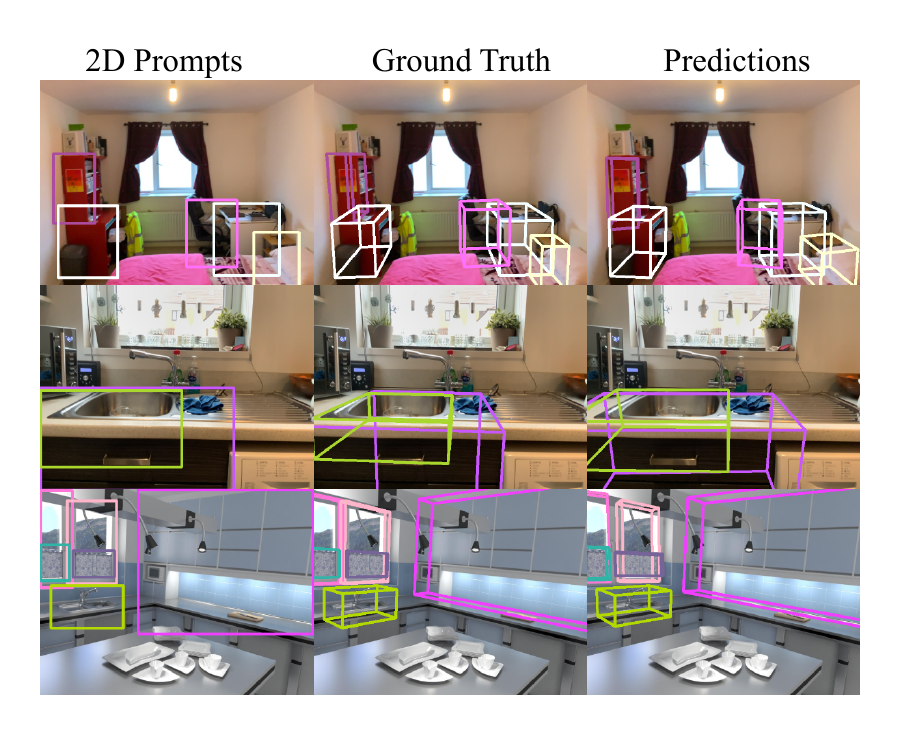}
    \caption{\textbf{Generalisation Performance: Results for Cross-Dataset Test.} 
    We show predictions on ARKitScenes and Hypersim made by our method trained on SUN RGB-D.}%
    \label{fig: cross-data}
\end{figure}

\boldstart{Total3DUnderstanding~\cite{nie2020total3dunderstanding}} We use their publicly released code and the model pre-trained on SUN RGB-D in experiments of Section 5.

\boldstart{Cube R-CNN~\cite{brazil2023omni3d}} For the results on the SUN RGB-D dataset in the second row of Table 1 in the main paper, we use the numbers reported directly from their paper. For the other experiments in Section 5, we use their publicly available code and pre-trained models.
\begin{figure*}[t]
    \centering
    \includegraphics[width=0.92\linewidth]{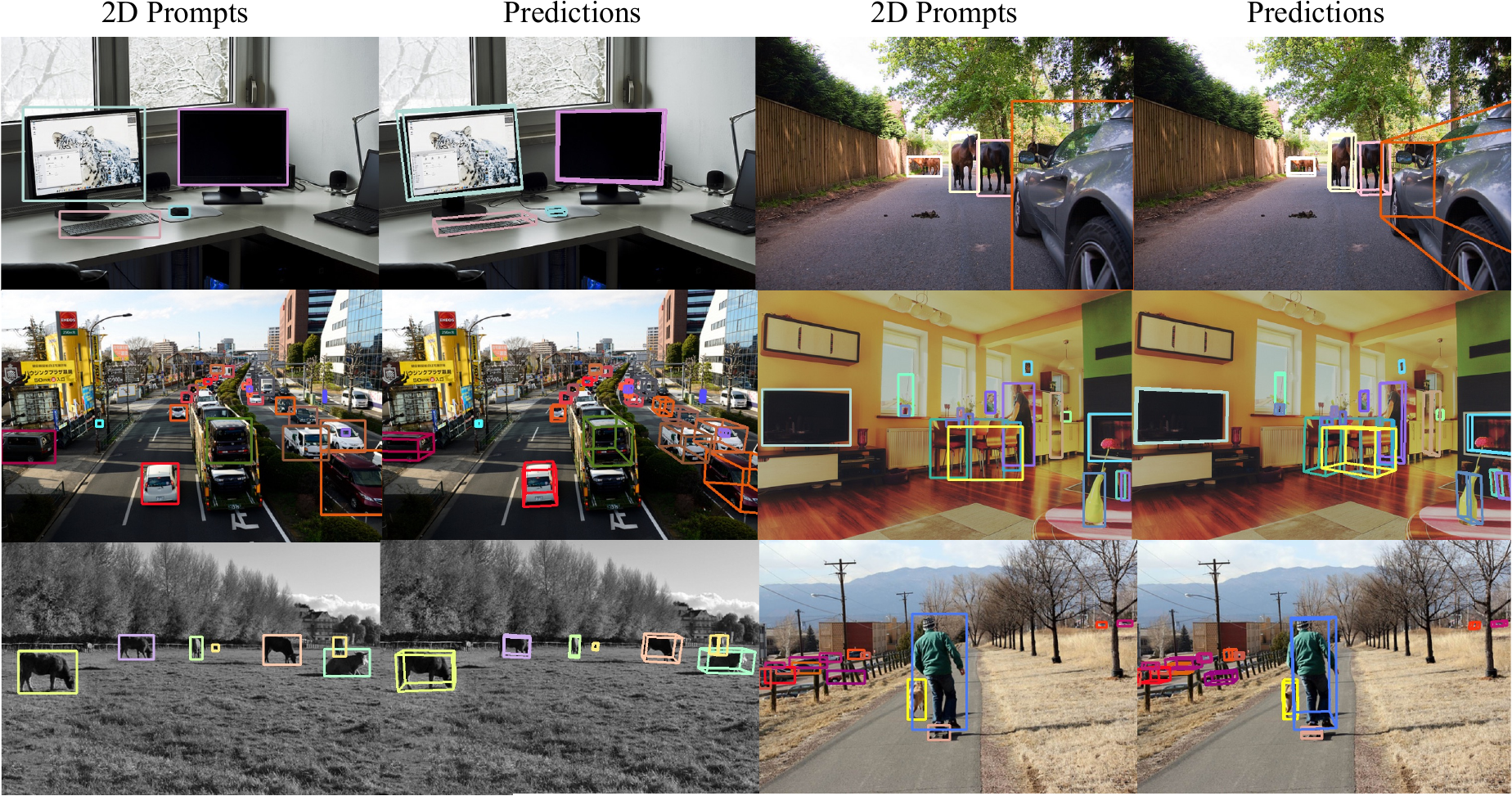}
    \caption{\textbf{Generalisation Performance: Additional Results for In-the-Wild Objects on COCO Dataset.} We show predictions made by our method without knowing object depths or camera intrinsics. By using constant values for depths and camera intrinsics, our approach accurately
predicts 3D boxes with well-aligned projections on the image.}%
    \label{fig: coco2}
\end{figure*}
\subsection{Algorithms}
The training and inference algorithms are shown in \cref{algorithm: training} and \cref{algorithm: sampling}.

\section{Additional Results}
\subsection{Generalisation}
\boldstart{Cross-dataset generalisation}
Our model trained on SUN RGB-D achieves an average IoU of 39.0 on the Hypersim~\cite{roberts2021hypersim} test set and 48.2 on the ARKitScenes~\cite{baruch2021arkitscenes} test set, highlighting its strong generalisation across different datasets. \cref{fig: cross-data} presents some of the test results.

\begin{table}[]
\caption{\textbf{Randomness Analysis on SUN RGB-D test set.} We evaluate the model using 10 different random seeds and report the mean, maximum, minimum, and standard deviation $\sigma$ for both IoU and NHD. }
\begin{tabular}{l|cccc}
    & Mean   & Max    & Min    & $\sigma$    \\ \hline
IoU (\%) $\uparrow$ & 61.38  & 61.46  & 61.24  & 6.4e-4 \\
NHD $\downarrow$  & 0.1140 & 0.1146 & 0.1133 & 3.9e-4
\end{tabular}
\label{table: rand}
\end{table}

\boldstart{Additional COCO results}
\cref{fig: coco2} shows additional results on COCO dataset.

\subsection{Per-category SUN RGB-D Performance}
In Table 1 of the main paper, we report the average IoU and NHD on 10 common categories of the SUN RGB-D dataset to make a fair comparison for baseline methods with different categories. \cref{table: cat1} and \cref{table: cat2} show the per-category IoU and NHD performances respectively.

\begin{table*}[]
\caption{\textbf{Per-category IoU (\%) on SUN RGB-D test set.} The top
three rows use GT 2D boxes along with predicted depths. The depths of our predictions are set to the same as \cite{brazil2023omni3d} for fair comparison. The bottom three rows use GT 2D boxes and GT depths for all methods.}
\resizebox{\textwidth}{!}{%
\begin{tabular}{l|l|llllllllll|l}
Methods     & Trained on & table & bed  & sofa & bathtub & sink & shelves & cabinet & fridge & chair & toilet & avg. \\ \hline
Total3D     & SUN RGB-D  & 28.0           & 37.0           & 30.1           & 27.6           & 20.1           & 10.8           & 14.3           & 20.2           & 24.8           & 35.4           & 24.8    \\
Cube R-CNN  & SUN RGB-D & 39.2          & 49.5          & 46.0          & 32.2          & 31.9          & 16.2          & 26.5          & 34.7          & 39.9          & 45.7          & 36.2       \\
Cube R-CNN  & Omni3D Indoor & 41.4  & 50.9 & \textbf{50.8} & \textbf{39.2}    & 35.0 & 17.8    & 28.2    & \textbf{35.1}   & 41.3  & 48.1   & 38.8     \\
Ours-d    & SUN RGB-D  & \textbf{42.2}          & \textbf{54.4}          & 50.5          & 38.9          & \textbf{40.3}          & \textbf{19.7}          & \textbf{29.4}          & 33.5          & \textbf{43.2}          & \textbf{50.1}          & \textbf{40.2}          \\ \hline
Total3D*    & SUN RGB-D  & 45.0           & 47.9           & 49.7           & 49.5           & 44.8           & 30.8           & 38.2           & 48.2           & 56.3           & 55.8           & 46.6     \\
Cube R-CNN* & Omni3D Indoor & 54.8  & 57.0 & 62.9 & 52.7    & 49.7 & 37.5    & 47.6    & 58.5   & 63.6  & 61.7   & 54.5     \\
Ours*        & SUN RGB-D  & \textbf{63.1} & \textbf{64.3} & \textbf{64.8} & \textbf{56.7} & \textbf{62.6} & \textbf{44.0} & \textbf{56.5} & \textbf{62.2} & \textbf{70.3} & \textbf{68.9} & \textbf{61.4} \\
\end{tabular}
}

\label{table: cat1}
\end{table*}

\begin{table*}[t]
\caption{\textbf{Per-category NHD on SUN RGB-D test set.} The top
three rows use GT 2D boxes along with predicted depths. The depths of our predictions are set to the same as \cite{brazil2023omni3d} for fair comparison. The bottom three rows use GT 2D boxes and GT depths for all methods.}
\resizebox{\textwidth}{!}{%
\begin{tabular}{l|l|llllllllll|l}
Methods     & Trained on & table          & bed            & sofa           & bathtub        & sink           & shelves        & cabinet        & fridge         & chair          & toilet         & avg.           \\ \hline
Total3D     & SUN RGB-D  & 0.352 & 0.254 & 0.314 & 0.288 & 0.526 & 0.497 & 0.443 & 0.380 & 0.408 & 0.297 & 0.376         \\
Cube R-CNN  & Omni3D Indoor & 0.230 & 0.162 & \textbf{0.164} & \textbf{0.215} & 0.244 & 0.324 & 0.384 & \textbf{0.229} & 0.233 & 0.172 & 0.236          \\
Ours-d      & SUN RGB-D  & \textbf{0.219} & \textbf{0.156} & 0.167 & 0.219 & \textbf{0.231} & \textbf{0.322} & \textbf{0.372} & 0.233 & \textbf{0.230} & \textbf{0.162} & \textbf{0.231} \\ \hline
Total3D*    & SUN RGB-D  & 0.204 & 0.168 & 0.180 & 0.157 & 0.188 & 0.251 & 0.210 & 0.177 & 0.148 & 0.160 & 0.184          \\
Cube R-CNN* & Omni3D Indoor & 0.148 & 0.125 & 0.107 & 0.149 & 0.146 & 0.181 & 0.176 & 0.117 & 0.107 & 0.112 & 0.137          \\
Ours*       & SUN RGB-D  & \textbf{0.114} & \textbf{0.107} & \textbf{0.101} & \textbf{0.120} & \textbf{0.114} & \textbf{0.161} & \textbf{0.127} & \textbf{0.111} & \textbf{0.093} & \textbf{0.090} & \textbf{0.114}
\end{tabular}
}

\label{table: cat2}
\end{table*}

\subsection{Randomness Analysis}
As discussed in Section 5.1 of the main paper, the diffusion process involves inherent randomness, so we conducted the experiments using 10 different random seeds and report the averaged results. To assess the model's stability, in addition to the averaged value reported in the main paper, we also provide the maximum, minimum, and standard deviation across these 10 runs in \cref{table: rand}.

\subsection{Noise on 2D Box}
We analyse the model's robustness towards noise during inference in \cref{table: noise}. We simulate box noise by applying Gaussian noise to box scales and translations separately, which can be written as:
\begin{equation}
\begin{aligned}
w' &= w + \mathcal{N}(0, \sigma_{\text{scale}}^2) \
\
h' &= h + \mathcal{N}(0, \sigma_{\text{scale}}^2)
\end{aligned},
\end{equation}

\begin{equation}
\begin{aligned}
x' &= x + \mathcal{N}(0, \sigma_{\text{trans}}^2 \cdot w) \
\
y' &= y + \mathcal{N}(0, \sigma_{\text{trans}}^2 \cdot h)
\end{aligned},
\end{equation}
where $w, h$ are the height and width of the ground truth boxes, $x, y$ are the centre coordinates and $w', h', x', y'$ are the noisy parameters. $\sigma_{\text{scale}}^2$ and $\sigma_{\text{trans}}^2$ are the variances of scale and translation noise. 
\cref{table: noise} shows that while the model is robust to noise in box scale and translation, translation errors have a greater impact on accuracy.

\begin{table}[]
\centering
\caption{\textbf{Noise on the 2D box}. We add different levels of random noise to the scale and translation of the 2D object box and report the model performance with these noisy box inputs.}
\begin{tabular}{ll|ll}
$\sigma_\text{scale}$ & $\sigma_\text{trans}$ & IoU (\%) $\uparrow$ & NHD $\downarrow$   \\ \hline
0.00       & 0.00       & 61.4  & 0.114 \\
0.05       & 0.00       & 59.9  & 0.120 \\
0.00       & 0.05       & 56.1  & 0.132 \\
0.05       & 0.05       & 55.2  & 0.135 \\
0.10       & 0.10       & 46.1  & 0.174
\end{tabular}
\label{table: noise}
\end{table}

\clearpage

\begin{algorithm}[t]
\caption{Training}
\begin{lstlisting}[language=Python]
def train_loss(images, gt_cubes, boxes_2d):
    """
    images: [B, H, W, 3]
    gt_cubes: [B, 1, D]
    boxes_2d: [B, 4]
    
    D: dimension of cubes
    N_train: number of sampled boxes during training
    """

    # Encode image features 
    feats = image_encoder(images)

    # Separate depth information 
    # from cube parameters
    cube_params, depths = separate_depth(gt_cubes)

    # normalise cube_params to [0, 1]
    cube_params = normalise_cube(cube_params)
    
    # Duplicate cube_params to N_train 
    x_0 = duplicate_cubes(cube_params) 
    
    # Signal scaling 
    x_0 = (x_0 * 2 - 1) * scale
    
    # Corrupt x_0
    t = randint(0, T) # time step 
    eps = normal(mean=0, std=1) # noise: [B, N_train, D-1]
    x_t = (
        sqrt(alpha_cumprod(t)) * x_0 
        + sqrt(1 - alpha_cumprod(t)) * eps
    )
    
    # Predict
    x_0_pred = denoising_model(
        x_t, feats, t, boxes_2d, depths
    )
    
    # Set prediction loss 
    loss = L(x_0_pred, gt_cubes)
    
    return loss
\end{lstlisting}
\label{algorithm: training}
\end{algorithm}

\begin{algorithm}[t]
\caption{Inference}
\begin{lstlisting}[language=Python]
def infer(images, steps, T, boxes_2d, depths):
    """
    images: [B, H, W, 3]
    steps: number of sampling steps
    T: total number of time steps
    boxes_2d: [B, 4]
    depths: object depths [B, 1]
    
    N_eval: number of proposal boxes during inference
    """
    
    # Encode image features
    feats = image_encoder(images)

    # Initialise noisy cube parameters (excluding depth) [B, N_eval, D-1]
    x_t = normal(mean=0, std=1)

    # Define uniform sampling step sizes
    times = reversed(
        linspace(0, T, steps)
    )

    # Generate pairs of consecutive time steps
    time_pairs = list(
        zip(times[:-1], times[1:])
    )

    # Iterate through time pairs
    for t_now, t_next in time_pairs:
        # Predict cube parameters x_0 from x_t
        x_0_pred = denoising_model(
            x_t, feats, t_now, boxes_2d, depths
        )

        # Estimate x_t at t_next
        x_t = ddim_step(
            x_t, x_0_pred, t_now, t_next
        )

    # Combine predicted cube parameters with depth information
    pred_cubes = combine_cubes(x_0_pred, depths)

    return pred_cubes
    
\end{lstlisting}
\label{algorithm: sampling}
\end{algorithm}

\end{document}